\documentclass[10pt,twocolumn,letterpaper]{article}

\usepackage[pagenumbers]{cvpr} %

\usepackage{amsmath,amsfonts,bm}

\def\eqref#1{equation~\ref{#1}}

\def\1{\bm{1}}

\def\vmu{{\bm{\mu}}}
\def\vtheta{{\bm{\theta}}}

\def\vepsilon{{\bm{\epsilon}}}

\def\vb{{\bm{b}}}
\def\vc{{\bm{c}}}

\def\vx{{\bm{x}}}

\def\mI{{\bm{I}}}

\DeclareMathAlphabet{\mathsfit}{\encodingdefault}{\sfdefault}{m}{sl}
\SetMathAlphabet{\mathsfit}{bold}{\encodingdefault}{\sfdefault}{bx}{n}

\def\gN{{\mathcal{N}}}

\def\sR{{\mathbb{R}}}

\newcommand{\E}{\mathbb{E}}

\usepackage{graphicx}
\usepackage{amsmath}
\usepackage{amssymb}
\usepackage{booktabs}
\usepackage{float}
\usepackage{makecell}
\usepackage{graphicx}
\usepackage{multirow}
\usepackage{colortbl}
\usepackage{xcolor}

\usepackage[pagebackref,breaklinks,colorlinks]{hyperref}

\usepackage[capitalize]{cleveref}
\crefname{section}{Sec.}{Secs.}
\Crefname{section}{Section}{Sections}
\Crefname{table}{Table}{Tables}
\crefname{table}{Tab.}{Tabs.}

\def\attntwoout{$\mathrm{CA}_{out}$}
\def\attntwocross{$\mathrm{CA}_{c}$}
\def\ffnin{$\mathrm{FFN}_{in}$}

\begin{document}

\title{A Closer Look at Parameter-Efficient Tuning in Diffusion Models}

\author{
Chendong Xiang$^1$, Fan Bao$^1$, Chongxuan Li$^2$, Hang Su$^{1,3}$, Jun Zhu$^{1,3}$\thanks{Corresponding author.} \\
$^1$Dept. of Comp. Sci. \& Tech., Institute for AI, BNRist Center \\
$^1$Tsinghua-Bosch Joint ML Center, THBI Lab,Tsinghua University, Beijing, 100084 China \\
$^2$Gaoling School of Artificial Intelligence, Renmin University of China \\
$^2$Beijing Key Lab of Big Data Management \& Analysis Methods \quad   $^3$Pazhou Lab, Guangzhou \\
{\tt\small \{bf19,\ xcd19\}@mails.tsinghua.edu.cn;} \\
{\tt\small chongxuanli@ruc.edu.cn;\ \{suhangss,\ dcszj\}@tsinghua.edu.cn}
}
\maketitle

\begin{abstract}
Large-scale diffusion models like Stable Diffusion~\cite{stable-diffusion} are powerful and find various real-world applications while customizing such models by fine-tuning is both memory and time inefficient. Motivated by the recent progress in natural language processing, we investigate parameter-efficient tuning in large diffusion models by inserting small learnable modules (termed adapters). 
In particular, we decompose the design space of adapters into orthogonal factors -- the input position, the output position as well as the function form, and perform Analysis of Variance (ANOVA), a classical statistical approach for analyzing the correlation between discrete (design options) and continuous variables (evaluation metrics). 
Our analysis suggests that the input position of adapters is the critical factor influencing the performance of downstream tasks. Then, we carefully study the choice of the input position, and we find that putting the input position after the cross-attention block can lead to the best performance, validated by additional visualization analyses. Finally, we provide a recipe for parameter-efficient tuning in diffusion models, which is comparable if not superior to the fully fine-tuned baseline (e.g., DreamBooth) with only 0.75 \% extra parameters, across various customized tasks. Our code is available at~\small{\url{https://github.com/Xiang-cd/unet-finetune}}

\end{abstract}

\begin{figure}[t!]
\begin{center}
\subfloat[Tuned-parameters and CLIP similarity comparison between Our method with best setting and Dreambooth.]{\includegraphics[width=.95\columnwidth]{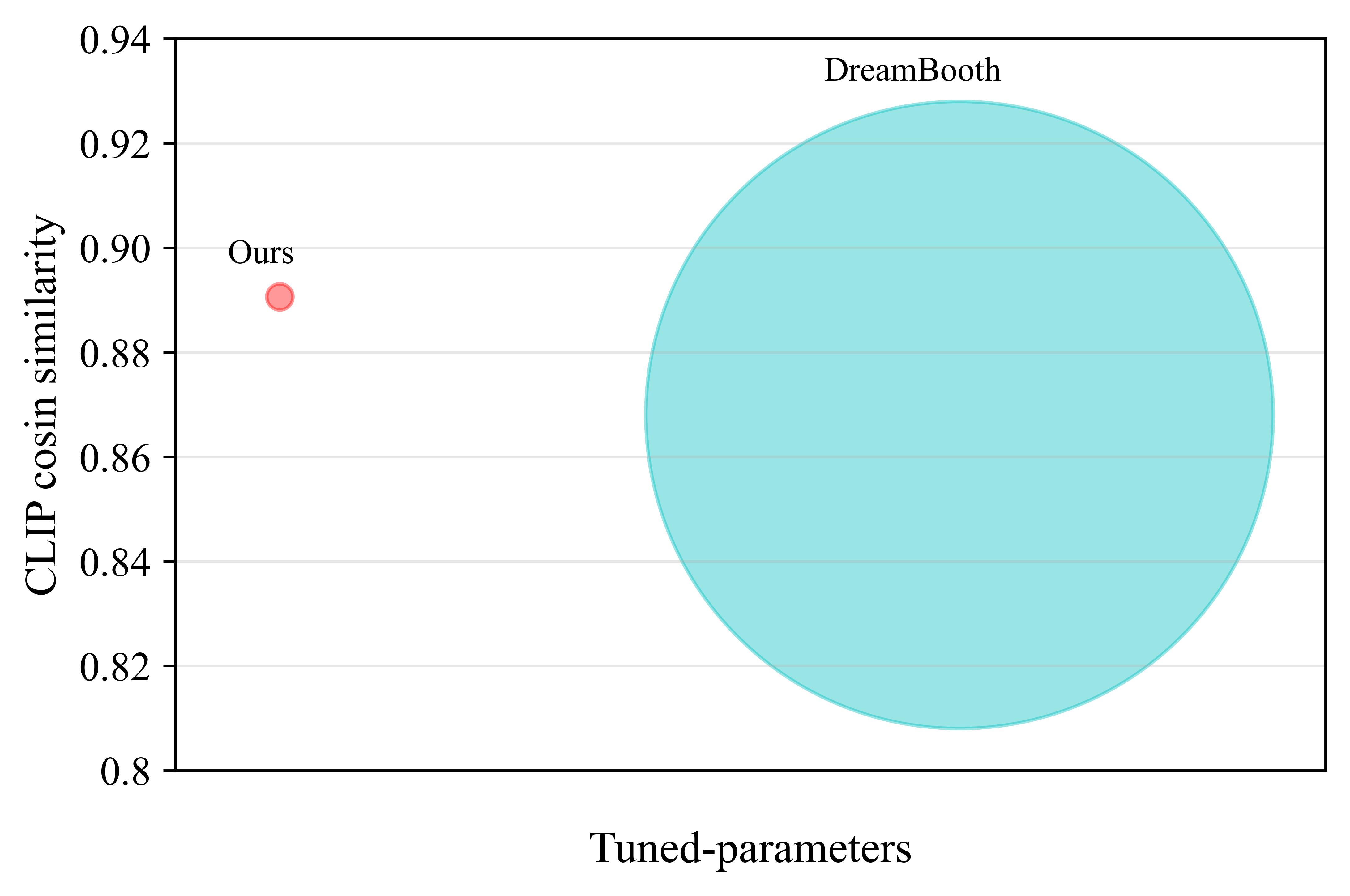}} \\ \vspace{.2cm}
\subfloat[Memory peak and time cost comparison between Our method with best setting and Dreambooth.]{\includegraphics[width=.95\columnwidth]{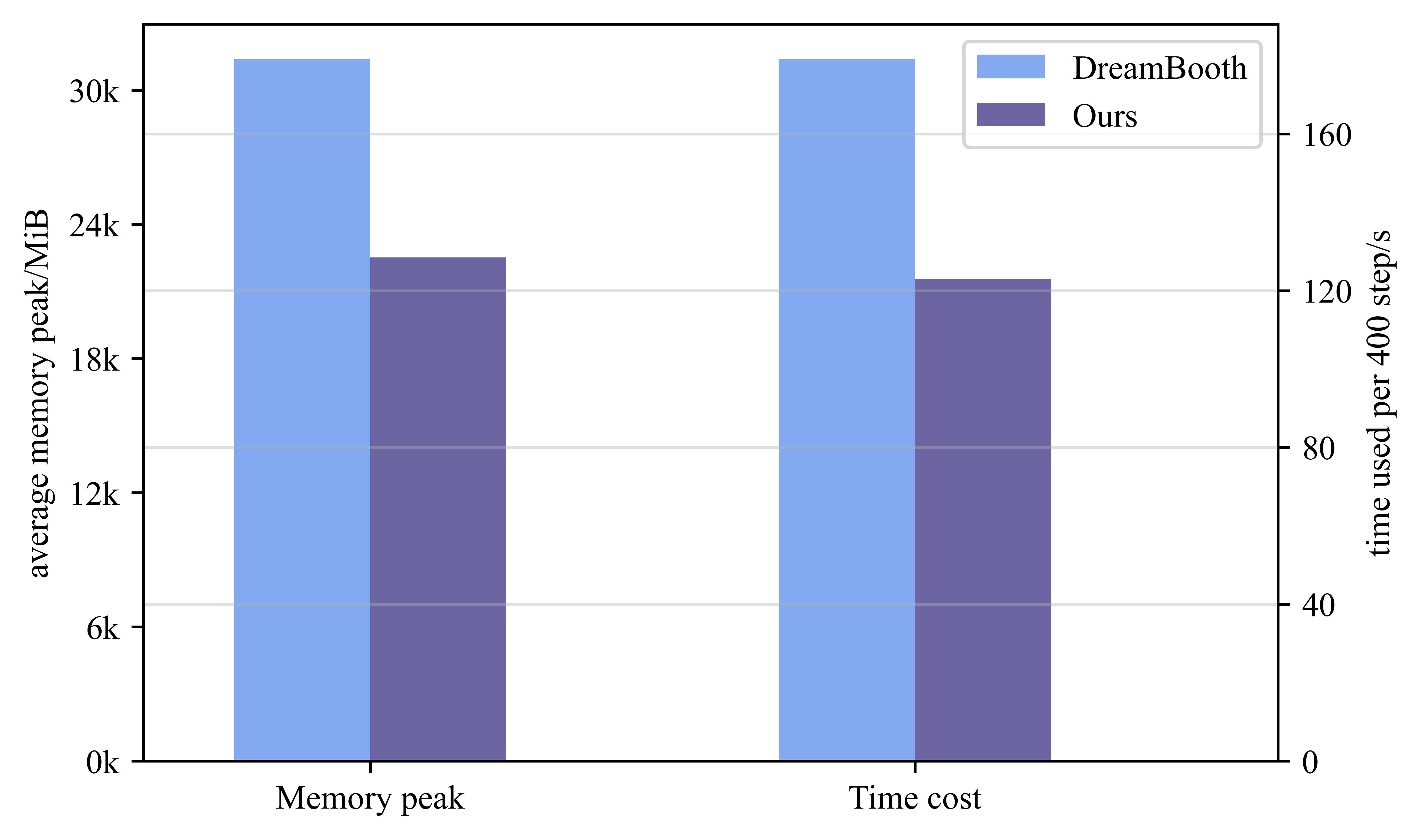}}
\end{center}
\caption{\textbf{Comparison of resource usage.} (a) Our method reaches comparable performance with much fewer parameters. (b) Our method reduces memory usage and time cost by around 30\%.}
\label{fig:accuracy}
\end{figure}

\begin{figure*}[t]
\centering
\includegraphics[width=\linewidth]{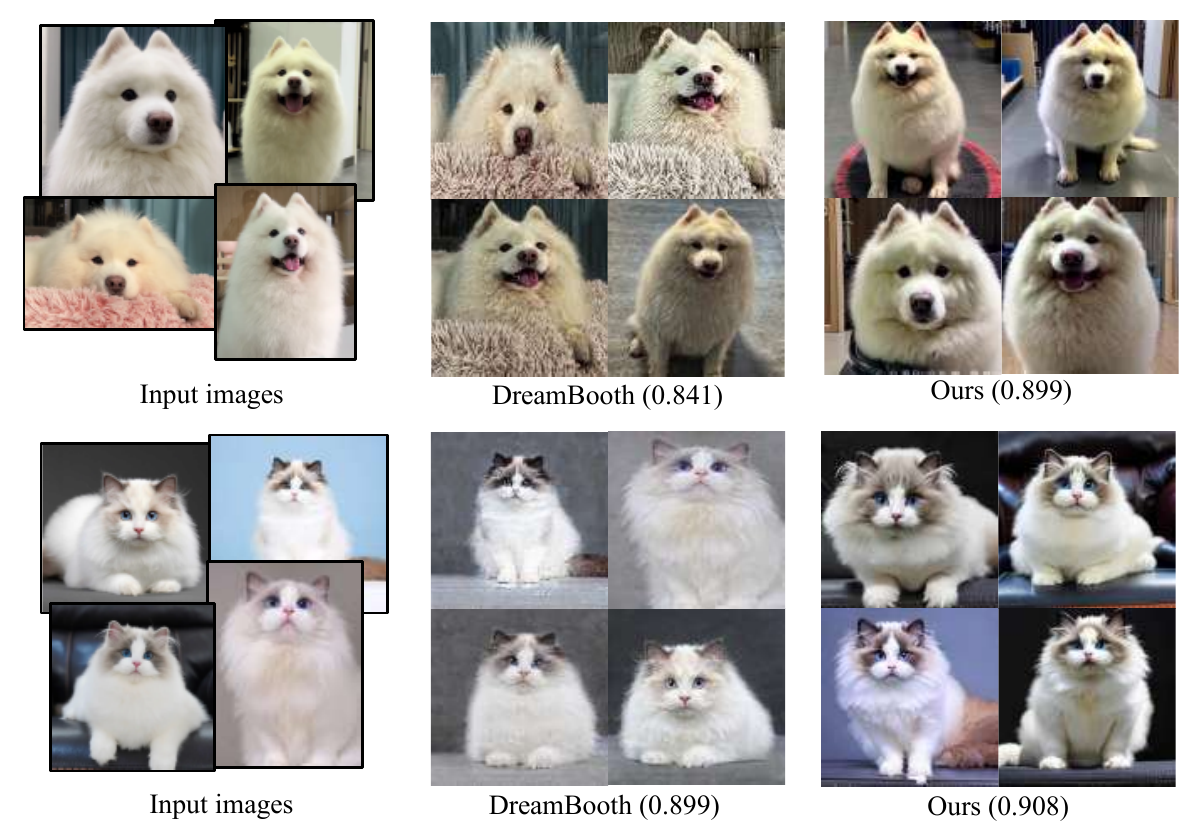}
\caption{\textbf{Comparison with DreamBooth.} Images generated by fully fine-tuned method (DreamBooth~\cite{ruiz2022dreambooth}) and our parameter-efficient tuning method with the best setting (Ours) on personalization tasks. We select the best samples for both methods (see more samples in Appendix). Ours achieves better performance in terms of both visual quality and the CLIP similarity$\uparrow$ (in brackets).}
\label{fig:first_page}
\end{figure*}

\section{Introduction}

Diffusion models~\cite{ddpm,song2020score,sohl2015deep} have recently become popular due to their excellent ability to generate high-quality and diverse images~\cite{dhariwal2021diffusion,dalle2, imagen, stable-diffusion}. 
By interacting with the condition information in its iterative generation process, diffusion models have an outstanding performance in conditional generation tasks, which motivate its applications such as text-to-image generation~\cite{dalle2, stable-diffusion, imagen}, image-to-image translation~\cite{choi2021ilvr,zhao2022egsde,meng2021sdedit}, image restoration~\cite{saharia2022image,kawar2022denoising}, 3D synthesis~\cite{dreamfusion}, audio synthesis~\cite{chen2020wavegrad,kong2020diffwave} and inverse molecular design~\cite{bao2022equivariant}.

With the knowledge learned from massive data, large-scale diffusion models act as strong priors for downstream tasks~\cite{dreamfusion,controlnet,ruiz2022dreambooth}. Among them, DreamBooth~\cite{ruiz2022dreambooth} tunes all parameters in a large-scale diffusion model to generate specific objects that users desire. However, fine-tuning the entire model is inefficient in terms of computation, memory and storage cost. %
An alternative way is the parameter-efficient transfer learning methods~\cite{houlsby2019parameter,he2022towards} originating from the area of natural language processing (NLP). These methods insert small trainable modules (termed as adapters) into the model and freeze the original model. Nevertheless, parameter-efficient transfer learning has not been thoroughly studied in the area of diffusion models. In contrast to the transformer-based language models~\cite{palm, gpt3, t5, bert} in NLP, the U-Net architecture widely used in diffusion models includes more components such as residual block with down/up-sampling operators, self-attention and cross-attention. This leads to a larger design space of parameter-efficient transfer learning than the transformer-based language models.

In this paper, we present a first systematical study on the design space of parameter-efficient tuning in large-scale diffusion models. We consider Stable Diffusion~\cite{stable-diffusion} as the concrete case, since currently it is the only open-source large-scale diffusion model.
In particular, we decompose the design space of adapters into orthogonal factors -- the input position, the output position, and the function form.
Through performing a powerful tool for analyzing differences between groups in experimental research named \emph{Analysis of Variance} (ANOVA)~\cite{girden1992anova} on these factors, we find that the input position is the critical factor influencing the performance of downstream tasks. 
Then, we carefully study the choice of the input position, and we find that putting the input position after the cross attention block can maximally encourage the network to perceive the change in input prompt (see Figure~\ref{fig:diff_pre}), therefore leading to the best performance.

Built upon our study, our best setting could reach comparable if not better results with the fully fine-tuned method within 0.75\% extra parameters on both the personalization task introduced in Dreambooth~\cite{ruiz2022dreambooth} and the task of fine-tuning on  a small set of text-image pairs.

 \begin{figure*}[t]
\centering
\includegraphics[width=\linewidth]{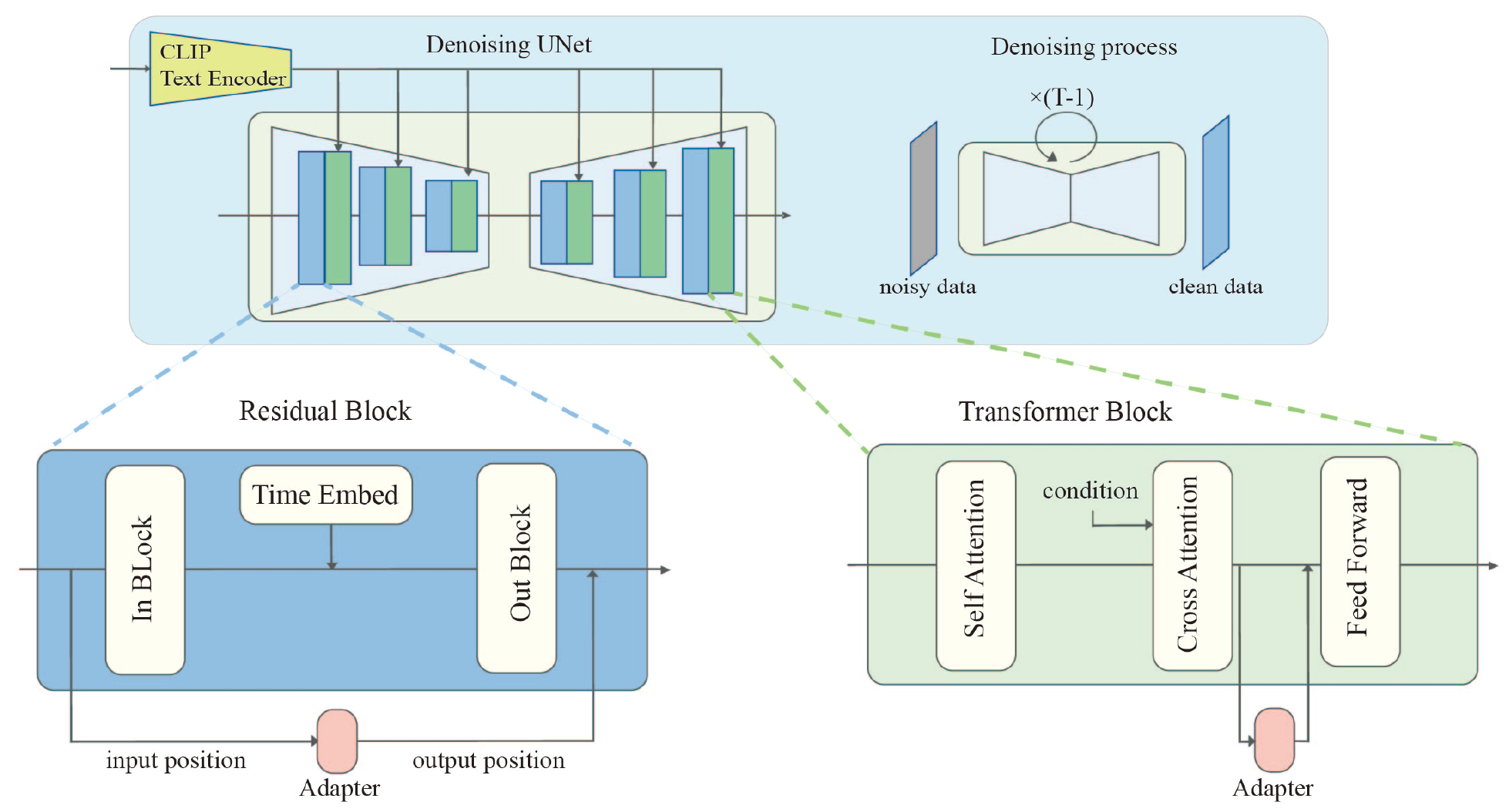}
\caption{\textbf{Background.} The top left figure shows the overview architecture of UNet-based diffusion model. The top right shows how the diffusion model removes noise from noisy data by $T-1$ steps. The bottom half of the figure shows the architecture of residual block and transformer block. Adapters (blocks with red color in the figure) are modules with a small number of parameters inserted into the model for parameter-efficient transfer learning. %
}
\label{fig:model arc}
\end{figure*}

\section{Background}

\subsection{Diffusion Models}

Diffusion models learn the data distribution $q(\vx_0)$ by reversing a noise-injection process 
\begin{align*}
    q(\vx_{1:T}|\vx_0) = \prod\limits_{t=0}^T q(\vx_t|\vx_{t-1}),
\end{align*}
where $q(\vx_t|\vx_{t-1}) = \gN(\vx_t|\sqrt{\alpha_t} \vx_{t-1}, \beta_t \mI)$ corresponds to a step of noise injection. The transition of the reverse is approximated by a Gaussian model $p(\vx_{t-1}|\vx_t) = \gN(\vx_{t-1}|\vmu(\vx_t), \sigma_t^2\mI)$, where the optimal mean under the maximal likelihood estimation~\cite{bao2022analytic} is
\begin{align*}
    \vmu_t^*(\vx_t) = \frac{1}{\sqrt{\alpha_t}} (\vx_t - \frac{\beta_t}{1-\overline{\alpha}_t} \E[\vepsilon|\vx_t]).
\end{align*}
Here $\overline{\alpha}_t = \prod_{i=1}^t \alpha_i$ and $\vepsilon$ is the standard Gaussian noise injected to $\vx_t$. To obtain the optimal mean, it is sufficient to estimate the conditional expectation $\E[\vepsilon|\vx_t]$ via a noise prediction objective
\begin{align*}
    \min_{\vtheta} \E_{t, \vx_0, \vepsilon} \| \vepsilon_\vtheta(\vx_t, t) - \vepsilon \|_2^2,
\end{align*}
where $\vepsilon_\vtheta(\vx_t, t)$ is the noise prediction network, and the optimal one satisfies $\vepsilon_{\vtheta^*}(\vx_t, t) = \E[\vepsilon|\vx_t]$ according to the property of $\ell_2$ loss.

In practice, we often care conditional generation. To perform it with diffusion models, we only need to introduce the condition information $c$ to the noise prediction network during training
\begin{align*}
    \min_{\vtheta} \E_{t, \vx_0, c, \vepsilon} \| \vepsilon_\vtheta(\vx_t, t, c) - \vepsilon \|_2^2.
\end{align*}

\subsection{The Architecture in Stable Diffusion}
Currently, the most popular architecture for diffusion models is the U-Net-based architecture~\cite{ddpm,dhariwal2021diffusion,dalle2,imagen,stable-diffusion}. Specifically, the U-Net-based architecture in Stable Diffusion~\cite{stable-diffusion} is shown in Figure~\ref{fig:model arc}. The U-Net comprises stacked basic blocks, each containing a transformer block and a residual block. 

In the transformer block, there are three types of sublayers: a self-attention layer, a cross attention layer, and a fully connected feed-forward network. 
The attention layer operates on queries $\boldsymbol{Q} \in \mathbb{R}^{n \times d_k}$, and key-value pairs $ \boldsymbol{K} \in \mathbb{R}^{m \times d_k}, \boldsymbol{V} \in \mathbb{R}^{m \times d_v}$:
\begin{align}
\label{eq:attention}
    \operatorname{Attn}(\boldsymbol{Q}, \boldsymbol{K}, \boldsymbol{V})    \in \mathbb{R}^{n \times d_v} =\operatorname{softmax}\left(\frac{\boldsymbol{Q} \boldsymbol{K}^T}{\sqrt{d_k}}\right) \boldsymbol{V} 
\end{align}
where $n$ is the number of queries, $m$ is the number of key-value pairs $d_k$ is dimension of key, $d_v$ is the dimension of value. In the self-attention layer, $\vx \in \mathbb{R}^{n \times d_x}$ is the only input. In the cross attention layer of conditioned diffusion model, there are two inputs $\vx \in \mathbb{R}^{n \times d_{x}} $ and $\vc \in \mathbb{R}^{m \times d_{c}}$, where $\vx$ is the output from prior block and $\vc$ represents the condition information.
The fully connected feed-forward network, which consists of two linear transformations with the ReLU activation function.:
\begin{align}
\label{eq:ffn}
    \operatorname{FFN}(\boldsymbol{x})=\operatorname{ReLU}\left(\boldsymbol{x} \boldsymbol{W}_1+\boldsymbol{b}_1\right) \boldsymbol{W}_2+\boldsymbol{b}_2
\end{align}
where $\boldsymbol{W}_1 \in \mathbb{R}^{d \times d_m}, \boldsymbol{W}_2 \in \mathbb{R}^{d_m \times d}$ are the learnable weights, and $\vb_1 \in \sR^{d_m}$, $\vb_2 \in \sR^d$ are the learnable biases. %

The residual block consists of a sequence of convolutional layers and activations, where the time embedding is injected into the residual block by an addition operation.

\subsection{Parameter-Efficient Transfer Learning}
Transfer learning is a technique that leverages the knowledge learned from one task to improve the performance of a related task. The method of pre-training and then performing transfer learning on downstream tasks is widely used. However, traditional transfer learning approaches require large amounts of parameters, which is computationally expensive and memory-intensive.

Parameter-efficient transfer learning is first proposed in the area of natural language processing (NLP). The key idea of parameter-efficient transfer learning is to reduce the number of updated parameters. This could be done by updating a part of the model or adding extra small modules.

Some parameter-efficient transfer learning methods (such as adapter~\cite{houlsby2019parameter}, LoRA~\cite{hu2022lora}) choose to add extra small modules named adapters to the model. In contrast, other methods (prefix tuning~\cite{li-liang-2021-prefix}, prompt-tuning~\cite{prompt-tuning}) prepend some learnable vectors to activations or inputs. Extensive study has validated that the efficient parameter fine-tuning method can achieve considerable results with a small number of parameters in the area of NLP.

\begin{figure}[t]
\centering
\includegraphics[width=\linewidth]{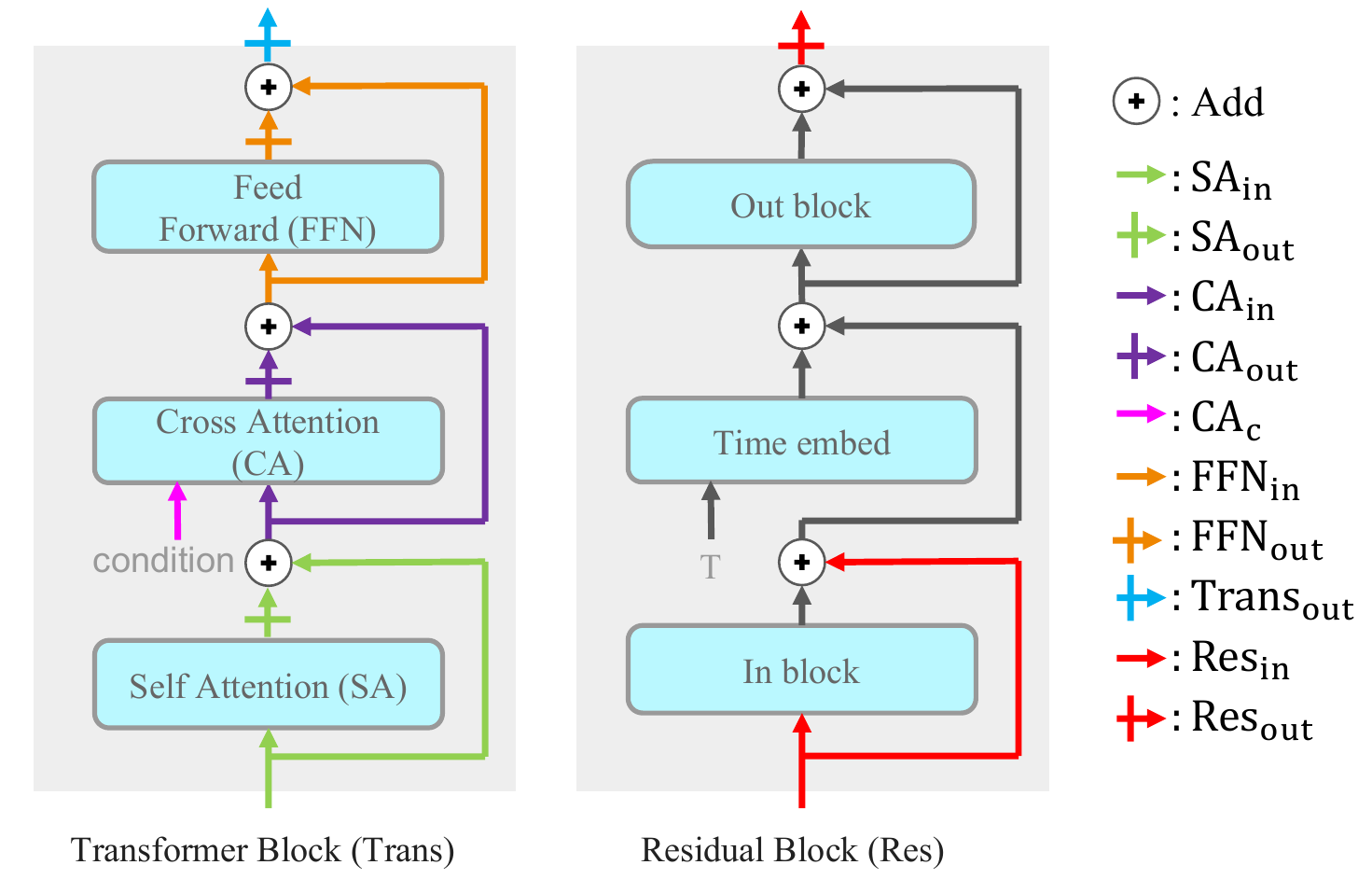}
    \caption{\textbf{Illustration of activation position.} generally, the main name of a activation position is an alias of a specific block in the model, the subscript of activation position explains the relationship between the activation and the block. 
    }
\label{fig:detail_data_path}
\end{figure}

\section{Design Space of Parameter-Efficient Learning in Diffusion Models}
\label{sec:design_space}

Despite the success of parameter-efficient transfer learning in NLP, this technique is not fully understood in the area of diffusion models due to the existence of more components such as the residual block and cross-attention. 

Before presenting our analysis on parameter-efficient tuning in diffusion models, we decompose the design space of adapters into three orthogonal factors -- the \textit{input position}, the \textit{output position}, and the \textit{function form}.

This work considers Stable Diffusion~\cite{stable-diffusion}, since currently it is the only open-source large-scale diffusion model (see Figure~\ref{fig:model arc} for its U-Net-based architecture).
Below we elaborate the input position, the output position, and the function form based on the architecture of Stable Diffusion.

\subsection{Input Position and Output Position}
\label{sec:inout}

The input position is where the adapter's input comes from, and the output position is where the adapter's output goes. 
For a neat notation, as shown in Figure~\ref{fig:detail_data_path}, the positions are named according to its neighboring layer. For example, $\mathrm{SA}_{in}$ represents that the position corresponds to the input of the self-attention layer, $\mathrm{Trans}_{out}$ corresponds to the output of the transformer block, and $\mathrm{CA}_c$ corresponds to the condition input of the cross attention layer.

In our framework, the input position could be any one of the activation positions described in Figure~\ref{fig:detail_data_path}. Thus, there are ten different options for the input position in total. As for output, some positions are equivalent since the addition is commutative. For example, putting output to $\mathrm{SA}_{out}$ is equivalent to putting output to $\mathrm{CA}_{in}$. As a result, the options for the output position are reduced to seven in total. Another constraint is that the output position must be placed after the input position.

\subsection{Function Form}
\begin{figure}[t]
\centering
\includegraphics[width=\linewidth]{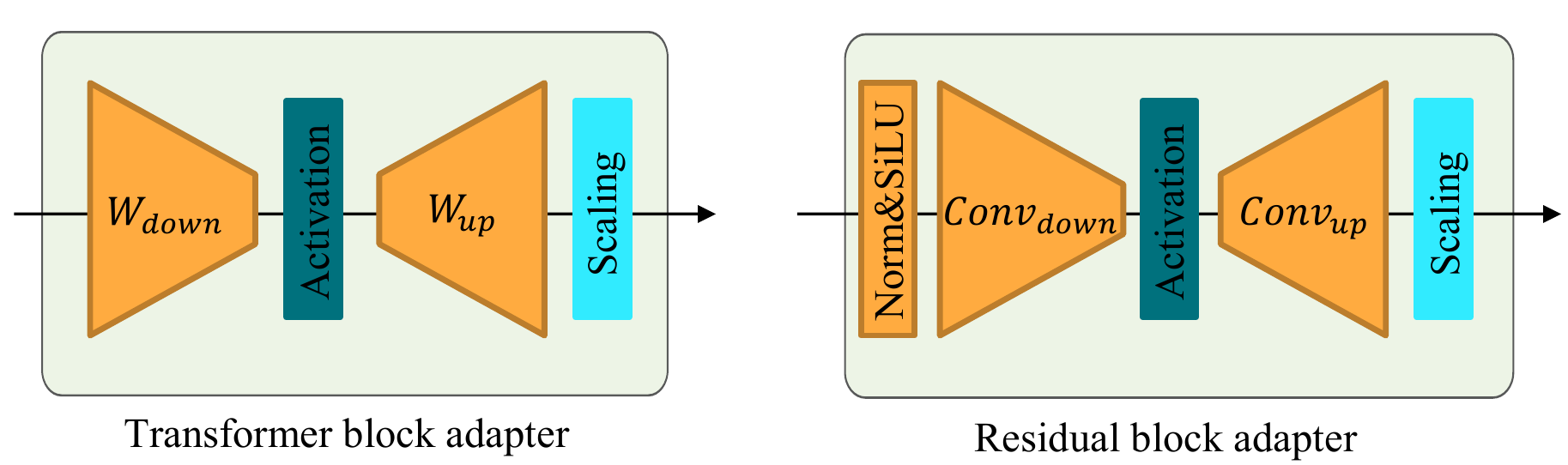}
\caption{The \textbf{function form} of adapters in the transformer block and residual block.}
\label{fig:func_form}
\end{figure}

Function form describes how an adapter transfers the input into the output. We present the function form of adapters in the transformer block and residual block respectively (see Figure~\ref{fig:func_form}), where both consist of a down-sampling operator, an activation function, an up-sampling operator, and a scaling factor. The down-sampling operator reduces the dimension of the input and the up-sampling operator increases the dimension to ensure the output has the same dimension as the input. The output is further multiplied with a scaling factor $s$ to control its strength in influencing the original network.

Specifically, the transformer block adapter uses low-rank matrices $W_{down}$ and $W_{up}$ as the down-sampling and up-sampling operators respectively, and the residual block adapter employs 3$\times$3 convolution layers $\mathrm{Conv}_{down}$ and $\mathrm{Conv}_{up}$ as the down-sampling and up-sampling operators respectively. Note that these convolution layers only change the number of channels without changing spatial size. Besides, the residual block adapter also processes its input with a group normalization~\cite{groupnorm} operator.

We include different activation functions and scaling factors in our design choice. 
The activation functions include $\mathrm{ReLU}$, $\mathrm{Sigmoid}$, $\mathrm{SiLU}$, and identity operator as our design choices, and the scale factors include 0.5, 1.0, 2.0, 4.0.

\section{Discover the Key Factor with Analysis of Variance}
\label{sec:anova}
As mentioned earlier, finding the optimal solution in such a large discrete search space is a challenge.
To discover which factor in the design space influences the performance the most, we quantify the correlation between model performance and factors by leveraging the one-way analysis of variance (ANOVA) method, which is widely used in many fields, including psychology, education, biology, and economics.

The main idea behind ANOVA is to partition the total variation in the data into two components: variation within groups (MSE) and variation between groups (MSB). MSB measures the difference between the group means, while the variation within groups measures the difference between individual observations and their respective group means.
The statistical test used in ANOVA is based on the F-distribution, which compares the ratio of the variation between groups to the variation within groups (F-statistic). If the F-statistic is large enough, it suggests that there is a significant difference between the means of the groups, which indicates a strong correlation.

\begin{figure}[t]
\centering
\includegraphics[width=\linewidth]{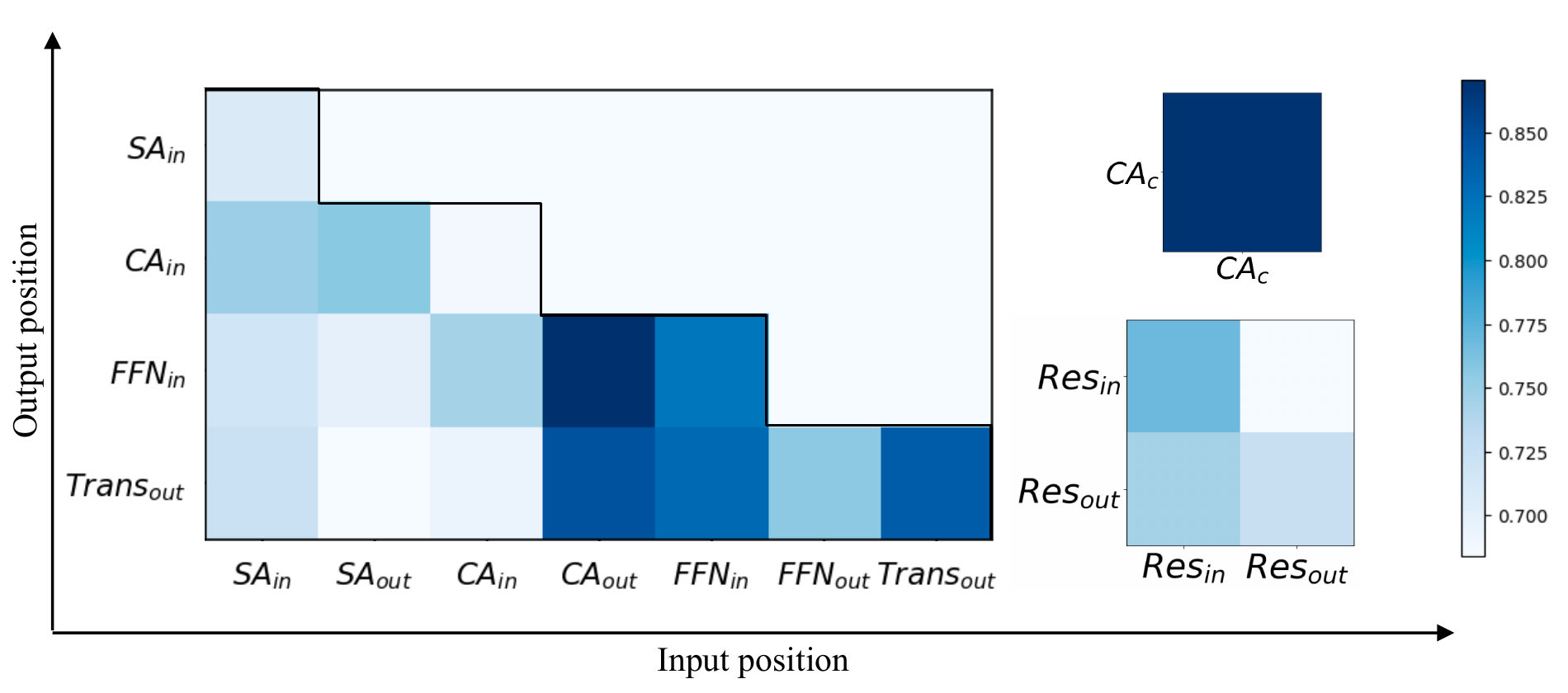}
\caption{The relationship between the performance (i.e., CLIP similarity$\uparrow$) and the input \& output position of adapters in the DreamBooth task.}
\label{fig:dreambooth_method_compare}
\includegraphics[width=\linewidth]{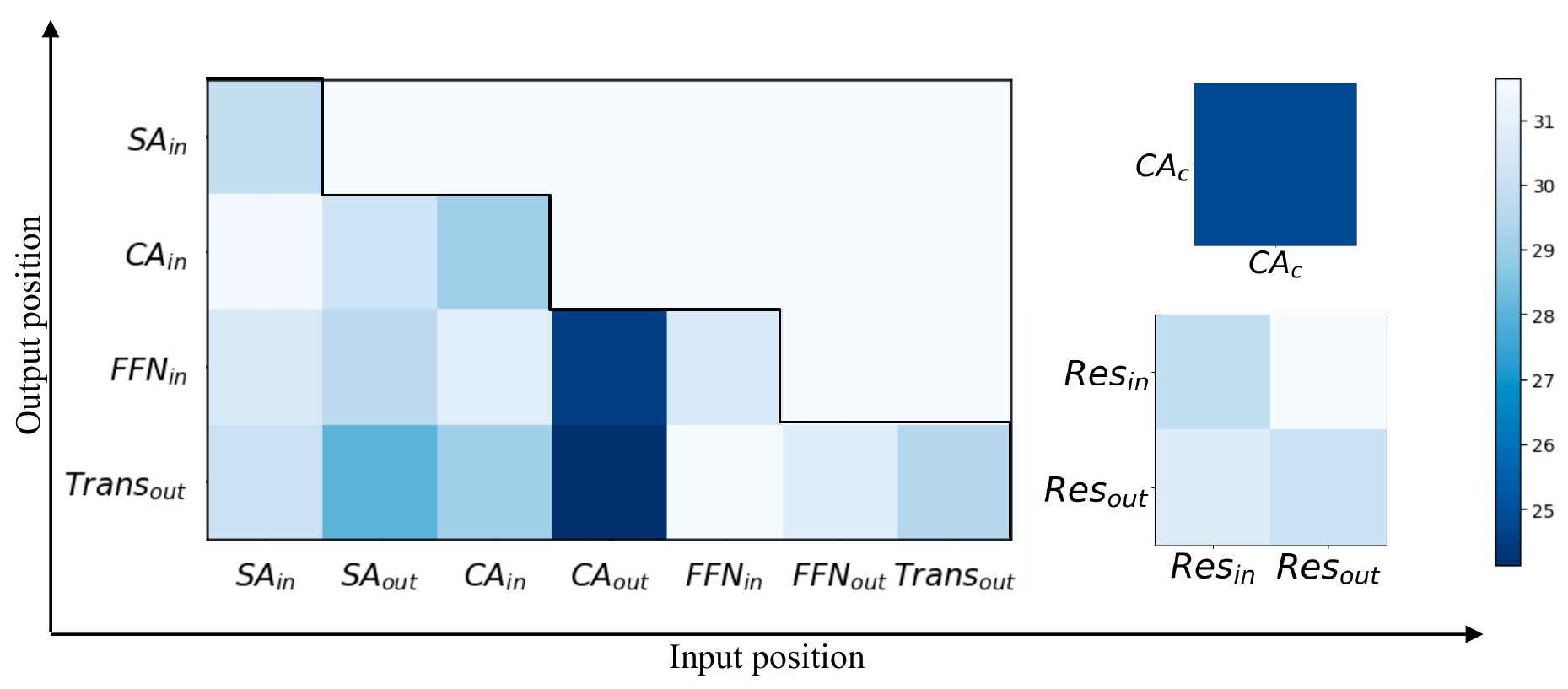}
\caption{The relationship between the performance (i.e., FID$\downarrow$) and the input \& output position of adapters in the fine-tuning task.}
\label{fig:flower_method_compare}
\end{figure}

\section{Experiments}
We first present our experimental setup in Section~\ref{sec:setup}. Then we analyze which factor in the design space is the most critical in Section~\ref{sec:design_space_exp}. After discovering the importance of the input position, we present a detailed ablation study on it in Section~\ref{sec:ablate}. Finally, we present a comprehensive comparison between our best setting and DreamBooth (i.e., fine-tuning all parameters) in Section~\ref{sec:cmp}.

\subsection{Setup}
\label{sec:setup}
\textbf{Tasks \& datasets}. 
We consider two transfer learning tasks in diffusion models characterized by different amount of data.

\textit{DreamBooth task.} The first task is to personalize diffusion models with less than 10 input images, as proposed in DreamBooth~\cite{ruiz2022dreambooth}. We term it DreamBooth task for simplicity.
The training dataset of DreamBooth consists of two sets of data: personalization data and regularization data. Personalization data is images of a specific object (e.g., a white dog) provided by the user. Regularization data is images of a general object similar to personalization data (e.g., dogs with different colors). Personalization data size is less than ten, and regularization data could be collected or generated by the model.
DreamBooth uses rare token $[V]$ and class word $C_{class}$ to distinguish regularization data and personalization data. In particular, with regularization data, the prompt will be ``a photo of $C_{class}$''; with personalization data, the prompt will be ``a photo of $[V]$ $C_{class}$''. Where $C_{class}$ is a word to describe the general class of data (e.g., dog). We collect personalization data from both the Internet and live-action photography, and also with data from DreamBooth (33 in total). we use Stable Diffusion itself to generate corresponding regularization data conditioned on prompt ``a photo of $C_{class}$''.

\textit{Fine-tuning task.} The other task is to fine-tune on a small set of text-image pairs. We term it fine-tuning task for simplicity.
Following \cite{transfer-diffusion}, we consider fine-tuning on flower dataset~\cite{flowerdataset} with 8189 images and use the same setting. We caption each image with the prompt ``a photo of $F_{name}$'', where $F_{name}$ is the flower name of the image class.

\textbf{Tuning}. %
We use AdamW~\cite{AdamW}  optimizer. For the DreamBooth task, we set the learning rate as 1e-4 which could let both DreamBooth and our method convergence around 1k step. fix the adapter size to 1.5M (0.17\% of the UNet model), and train with 2.5k steps. %
For the task of fine-tuning on a small set of text-image pairs, we set the learning rate as 1e-5, fix the adapter size to 6.4M (0.72\% of the UNet model), and train 60k steps.

\textbf{Sampling}. 
For a better sampling efficiency, we choose DPM-Solver~\cite{dpmsolver} as the sampling algorithm with 25 sampling steps, and a classifier free guidance (cfg)~\cite{ho2021classifier} scale of $7.0$. In some cases, we use cfg scale of $5.0$ for better image quality.

\textbf{Evaluation.}
For the DreamBooth task, we evaluate faithfulness using the image distance in the CLIP space as proposed in~\cite{gal2022textual}. Specifically, for each personalization target, we generate 32 images using the prompt: ``A photo of $[V] \ C_{class}$". The metric is the mean pair-wise CLIP-space cosine-similarity (CLIP similarity) between the generated images and the images of the personalization training set.

For the task on fine-tuning on a small set of text-image pairs, we use the FID score~\cite{FID} to evaluate the similarity between the training images and generated images. We randomly draw 5k prompts from the training set, use these prompts to generate images, and then compute FID by comparing generated images with training images.

\subsection{Analysis of Variance (ANOVA) on the Design Space}
\label{sec:design_space_exp}

Recall that we decompose the design space into factors of input position, output position, and function form. We perform ANOVA method (see Section~\ref{sec:anova} for details) on these design dimensions
We consider the DreamBooth task for efficiency, since it requires fewer training steps. 

\begin{figure}[t]
\centering
\includegraphics[width=\linewidth]{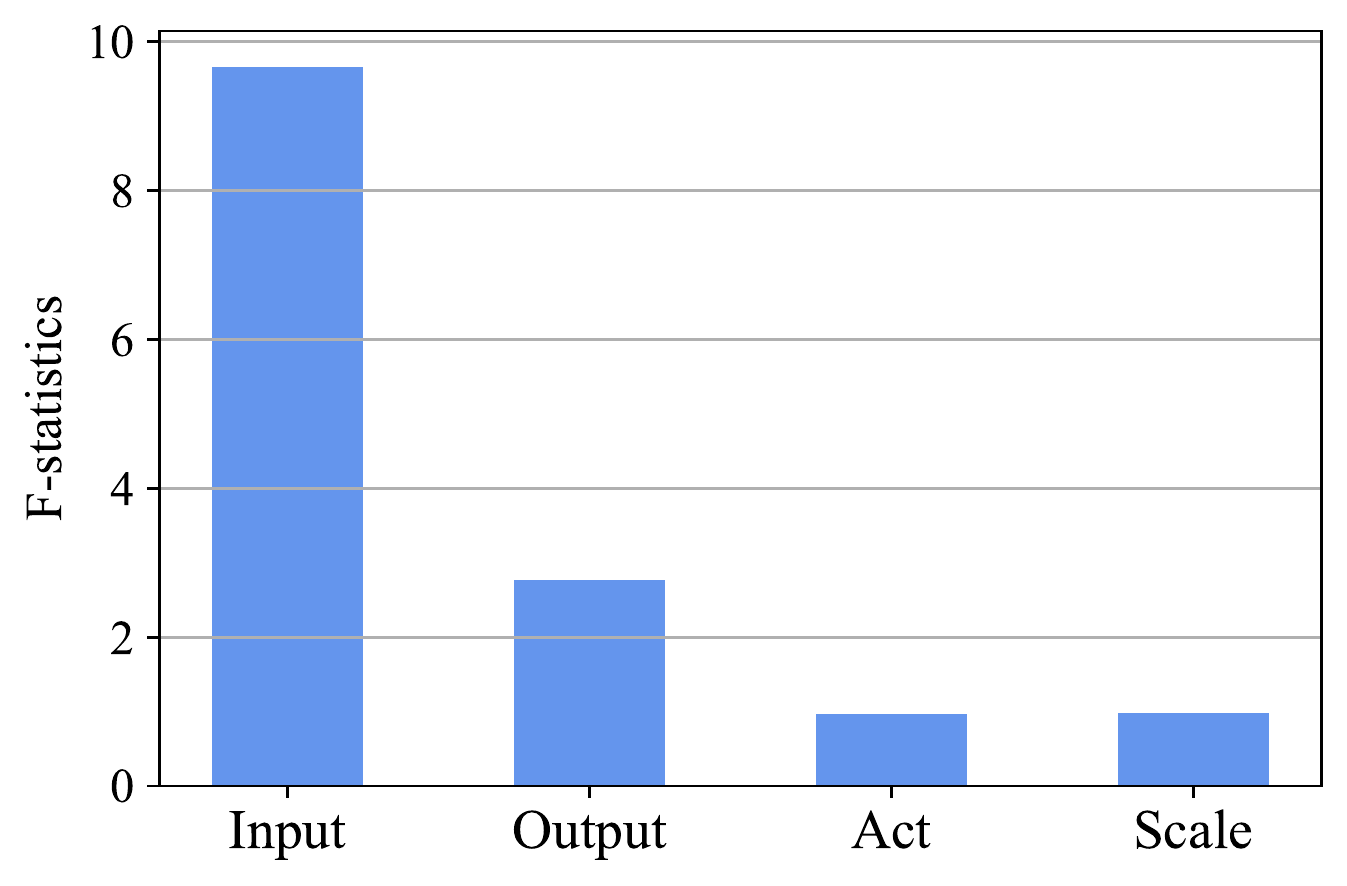}
\caption{F-statistic of ANOVA by grouping input position, output position, activation function, and scale factor. The F-statistic is large when grouping by input position, which indicates there is significant relation to the input position.}
\label{fig:f-sta}
\end{figure}

As shown in Figure~\ref{fig:f-sta}, when grouped by input position, the F-statistic is large, which indicates that the input position is a critical factor to model's performance. When grouped by output position, it shows a weak correlation. When grouped by function form (both activation function and scale factor), which have an F-statistic of around 1, indicating that the variability between groups is similar to the variability within groups, which suggests that there is no significant difference between the group means.

We further visualize the performance with different input positions and output positions.
Figure~\ref{fig:dreambooth_method_compare} shows the results of the DreamBooth task. 
Figure~\ref{fig:flower_method_compare} shows the FID results of the fine-tuning task. 

As discussed above, we conclude that the input position of adapter is the key factor affecting the performance of parameter-efficient transfer learning.

\begin{figure*}[t]
\centering
\includegraphics[width=\linewidth]{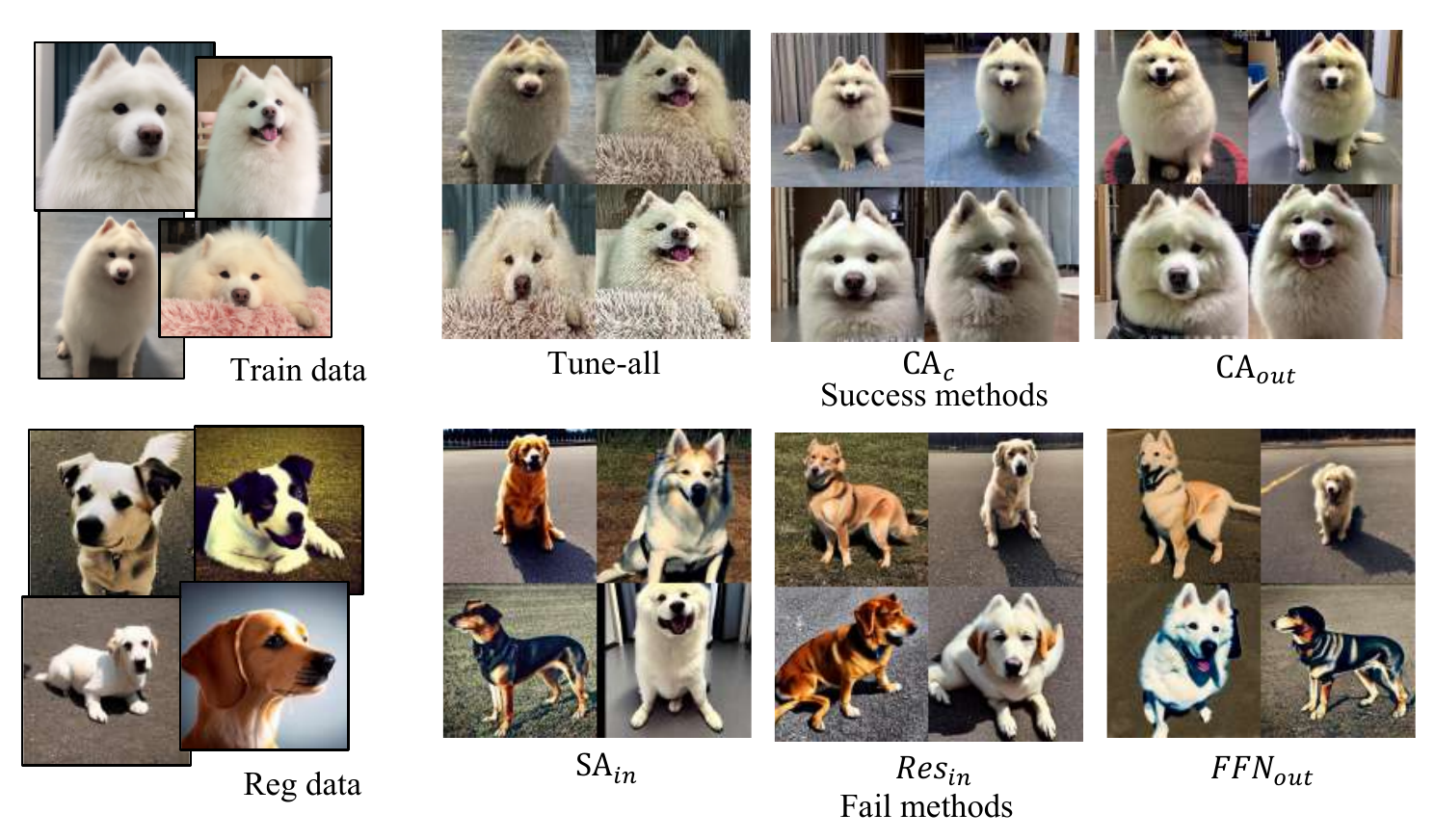}
\caption{\textbf{The generated samples of personalized diffusion models with different input positions of adapters.} All samples are conditioned on ``a photo of $[V]$ $C_{class}$'', it is worth noticing that the success methods generate the right images, but the fail methods are likely to generate pictures similar to regularization data.}
\label{fig:pic_compare}
\end{figure*}

\subsection{Ablate the Input Position}
\label{sec:ablate}

As shown in Figure~\ref{fig:dreambooth_method_compare} and Figure~\ref{fig:flower_method_compare}, we find that adapters with input position of \attntwocross{} or \attntwoout{} have a good performance on both tasks.
In Figure~\ref{fig:pic_compare}, we present generated samples in the personalized diffusion models with different input positions of adapters. Adapters with input position at \attntwocross{} or \attntwoout{} are able to generate personalized images comparable to fine-tuning all parameters, while adapters with input position at other places do not.

\begin{figure}[t]
\centering
\includegraphics[width=\linewidth]{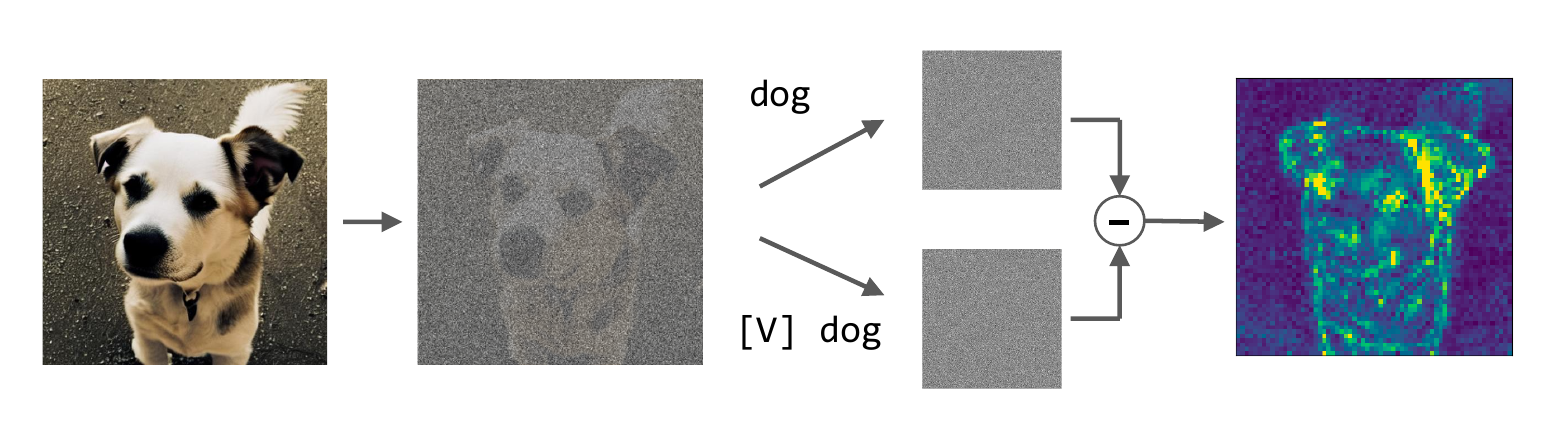}
\caption{Pipeline of experiment visualize the difference of noise prediction. }
\label{fig:pipeline}
\end{figure}

\begin{figure*}[t]
\includegraphics[width=\linewidth]{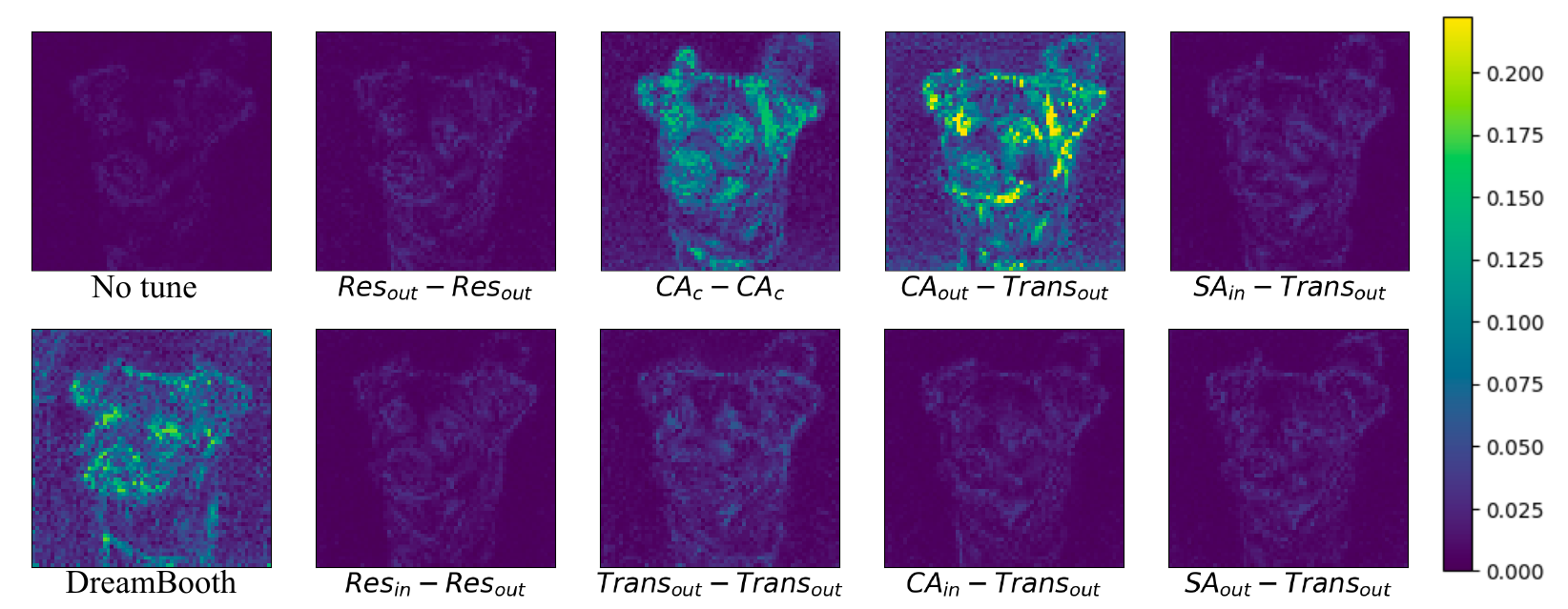}
\caption{\textbf{The noise prediction difference of various settings.} The ``No tune'' method uses the original Stable Diffusion model without any fine-tuning. All adapter methods are noted as the form of $input-ouput$. We found that adapters with input position of \attntwoout{} and \attntwocross{} react better with the prompt changes.}
\label{fig:diff_pre}
\end{figure*}

We further compute the difference between the noise prediction given prompt ``a photo of $[V]$ $C_{class}$" and ``a photo of $C_{class}$". The pipeline is shown in Figure~\ref{fig:pipeline}, where we firstly add noise to an image from the regularization data, use the U-Net to predict noise given the two prompts, and visualize the difference between the difference of two predicted noise. As shown in Figure~\ref{fig:diff_pre}, adapters with input position of \attntwocross{} or \attntwoout{} present a significant difference between the noise prediction.

\subsection{Compare with DreamBooth}
\label{sec:cmp}
In this section, we compare our best setting (with input position at \attntwoout{} and output position as \ffnin{}) to DreamBooth, which fine-tunes all parameters in diffusion models.

\begin{figure}[t]
\centering
\includegraphics[width=\linewidth]{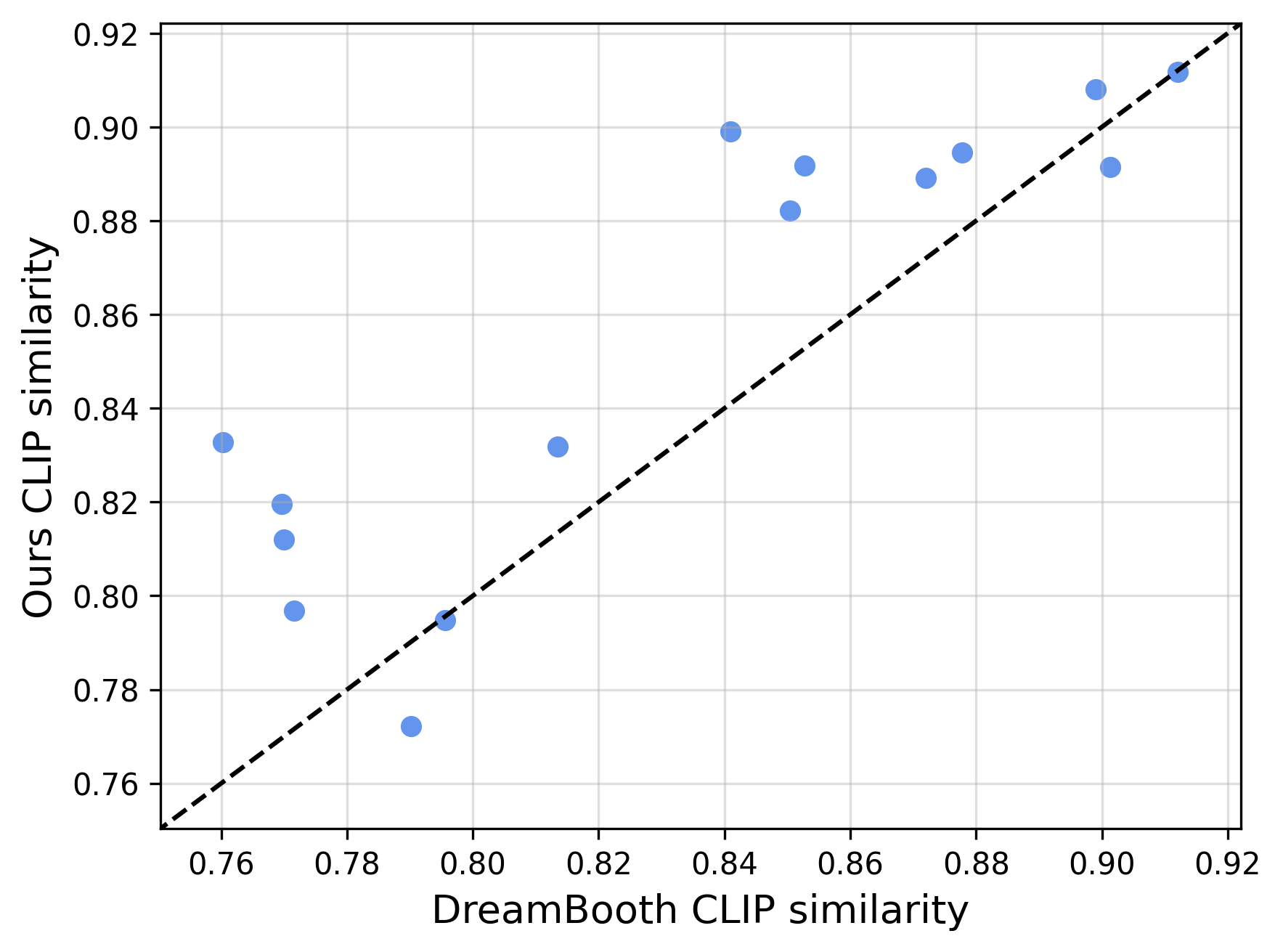}
\caption{Performance compare with DreamBooth. Our method performs better on most cases.}
\label{fig:scater}
\end{figure}

We show the results of each case on the DreamBooth task in Figure~\ref{fig:scater}, which show that our method is better in most cases.

We also compare our best setting to the fully fine-tuned method in the fine-tuning task on the flower dataset. Our recipe reach FID of 24.49, which is better than 28.15 of the fully fine-tuned method.

\section{Related Work}

\textbf{Personalization.} Large-scale text-to-image diffusion models trained on web data can generate high-resolution and diverse images whose contents are controlled by the input text, but often lacks the ability for personalized generation on a certain object that the user desires. Recent work such as textual inversion~\cite{gal2022textual} and DreamBooth~\cite{ruiz2022dreambooth} aims to address this by fine-tuning the diffusion model on a small set of images for the object. The textual inversion only tunes a word embedding. To obtain a stronger performance, DreamBooth tunes all parameters with a regularization loss to prevent overfitting.

\textbf{Parameter-efficient transfer learning}. Parameter-efficient transfer learning is originated from the area of NLP, such as adapter~\cite{houlsby2019parameter}, prefix tuning~\cite{li-liang-2021-prefix}, prompt tuning~\cite{prompt-tuning} and LoRA~\cite{hu2022lora}. Specifically, adapter~\cite{houlsby2019parameter} inserts small low-rank multilayer perceptron (MLP) with nonlinear activation function $f(\cdot)$ between transformer block; prefix tuning~\cite{li-liang-2021-prefix} prepends tunable prefix vectors to the keys and values at each attention layer; prompt-tuning~\cite{prompt-tuning} simplifies prefix-tuning by adding tunable input word embeddings; LoRA~\cite{hu2022lora} injects tunable low-rank matrices into the query and value projection matrices of the transformer block.

While these parameter-efficient transfer learning methods have different forms or motivations, recent work~\cite{he2022towards} proposes a unified view of these methods by designating a set of factors to describe the design space of parameter-efficient transfer learning in pure transformers~\cite{vaswani2017attention}. These factors include modified representation, insertion form, functional form, and composition function. 
In contrast, our method focuses on U-Net with more components than pure transformers, leading to a larger design space. Besides, we use a simpler way to decompose the design space into orthogonal factors, i.e., the input position, the output position and the function form. %

\textbf{Transfer learning for diffusion models}. 
There are methods that transfer the diffusion model to recognize a specific object or perform semantic editting~\cite{Imagic, ruiz2022dreambooth} by tuning the whole model. Previous work~\cite{transfer-diffusion} tries to transfer a large diffusion model into an image-to-image model on small datasets,  but the total number of parameters tuned is nearly half of the original model. \cite{T2I-Adapter, controlnet} transfers diffusion model to accept new conditions and introduces much more parameters than ours. Concurrent work~\cite{difflora} also performs parameter-efficient transfer learning on Stable Diffusion, their method could reach comparable results with fully fine-tuned method on DreamBooth~\cite{ruiz2022dreambooth} task,  while their method is based on adding adapters on multiple positions at the same time, leading to a more complicated design space.

\section{Conclusion}

In this paper, we perform a systematical study on the design space of parameter-efficient transfer learning by inserting adapters in diffusion models. We decompose the design space of adapters into orthogonal factors -- the input position, the output position and the function form.
By performing Analysis of Variance (ANOVA), we discover the input position of adapters is the critical factor influencing the performance of downstream tasks. 
Then, we carefully study the choice of the input position, and we find that putting the input position after the cross-attention block can lead to the best performance, validated by additional visualization analyses. Finally, we provide a recipe for parameter-efficient tuning in diffusion models, which is comparable if not superior to the fully fine-tuned baseline (e.g., DreamBooth) with only 0.75 \% extra parameters, across various customized tasks.

{\small
\bibliographystyle{ieee_fullname}
\bibliography{ egbib }
}

\appendix

\onecolumn

 \begin{figure}[H]
\centering
\includegraphics[width=\linewidth]{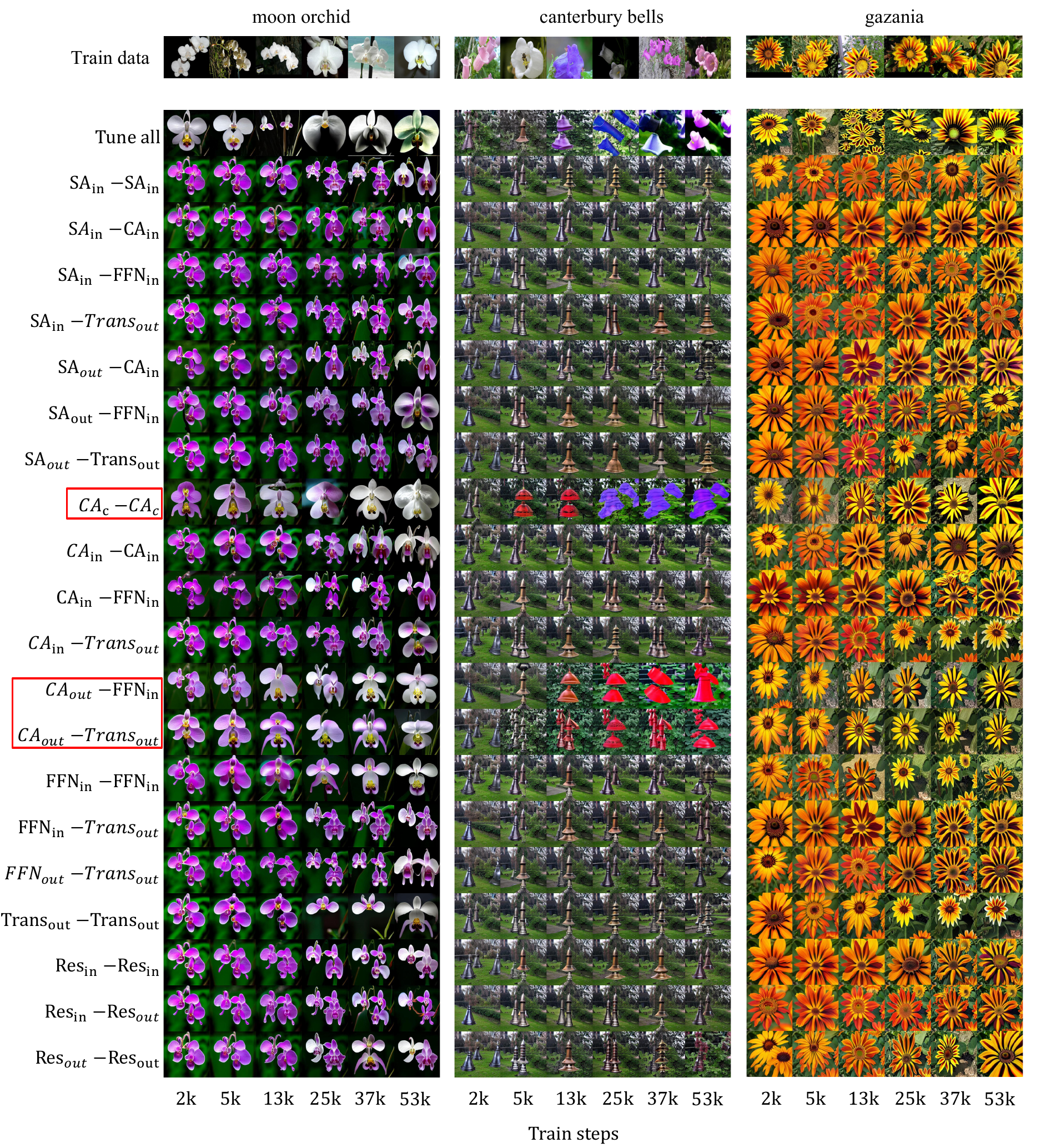}
\caption{\textbf{Example of fine-tuning on flower dataset.} We circle the method of reacting better with the prompt changes with red boxes. All examples are generated with the prompt ``a photo of $Class$.'' Where $Class$ is the class label of the flower (shown on the top). The left shows the case in which only a part of the method successfully turns the flower color into white. The middle shows the fail case that almost all methods failed to generate reasonable flowers due to the strong prior of the word ``bell''. The right shows the success case that almost every method works due to the strong prior of the origin model. 
}
\label{fig:flower example}
\end{figure}

\begin{figure}[H]
\centering
\includegraphics[width=\linewidth]{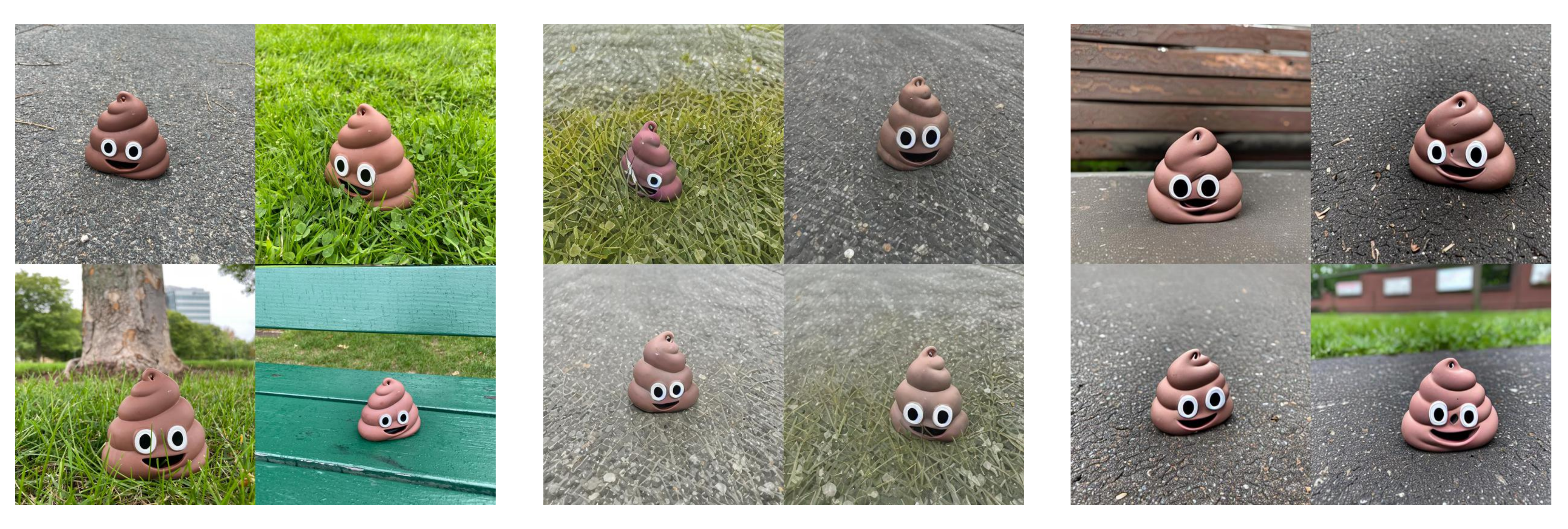}

\vspace{0.3cm}

\includegraphics[width=\linewidth]{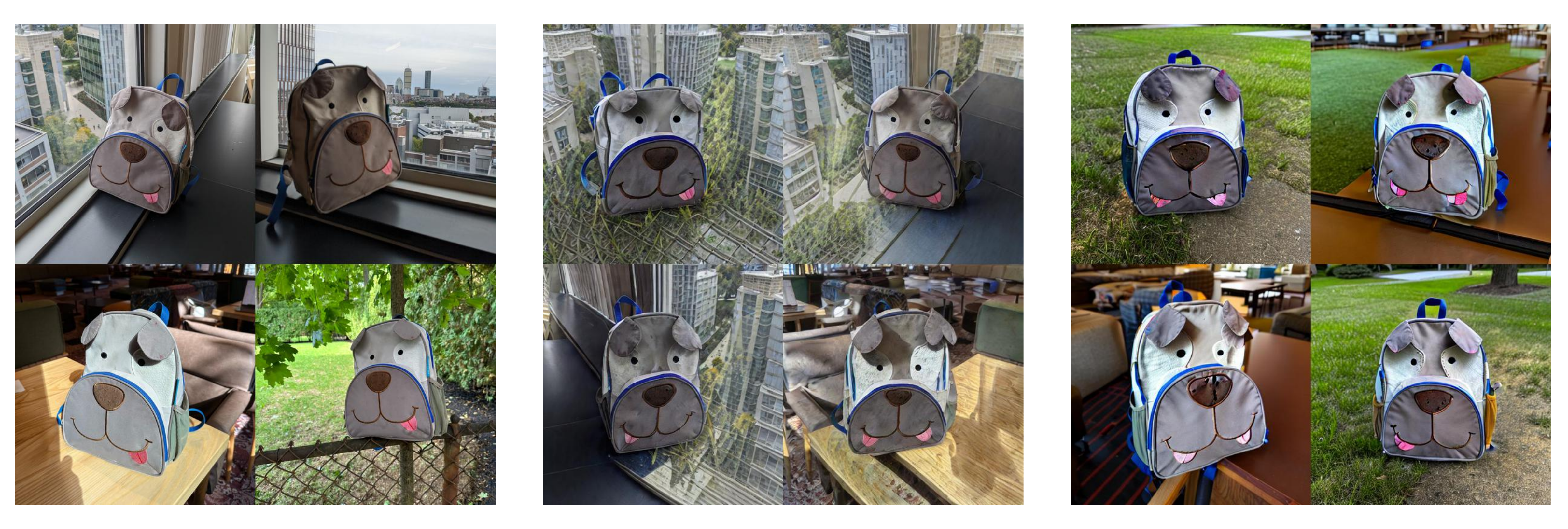}

\vspace{0.3cm}

\includegraphics[width=\linewidth]{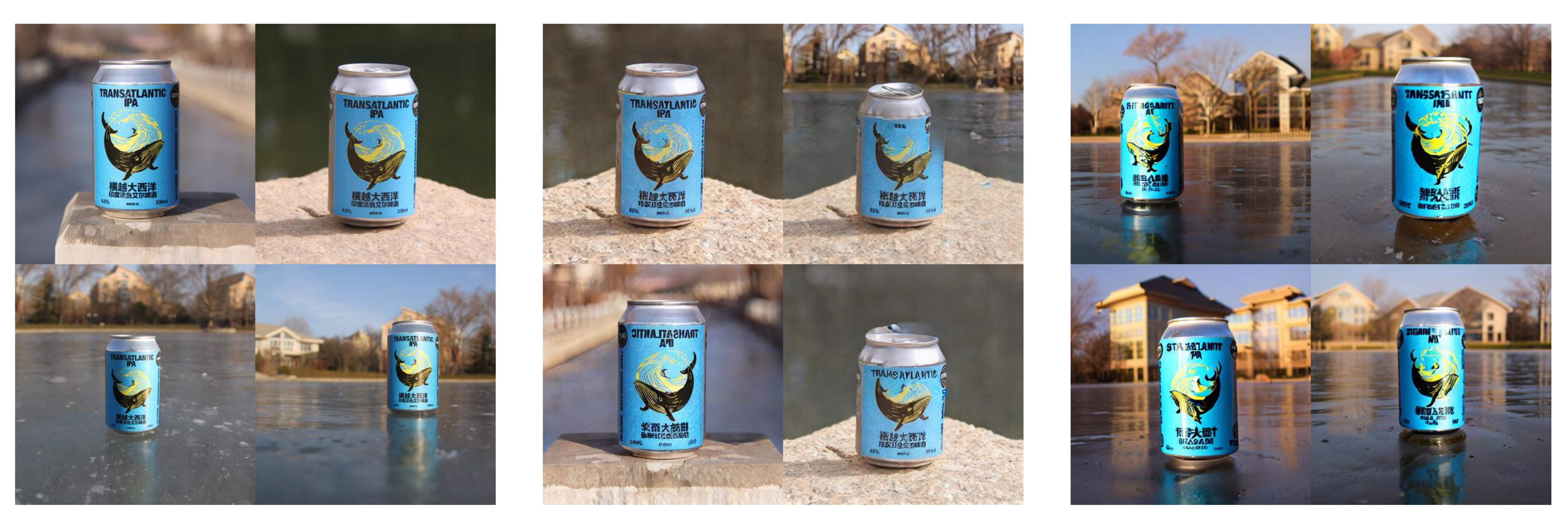}

\vspace{0.3cm}

\caption{\textbf{Example of DreamBooth}, train data (left), DreamBooth (middle), our method (right). All samples are randomly selected with the best training steps.}
\label{fig:dream example1}
\end{figure}

 \begin{figure}[H]
\includegraphics[width=\linewidth]{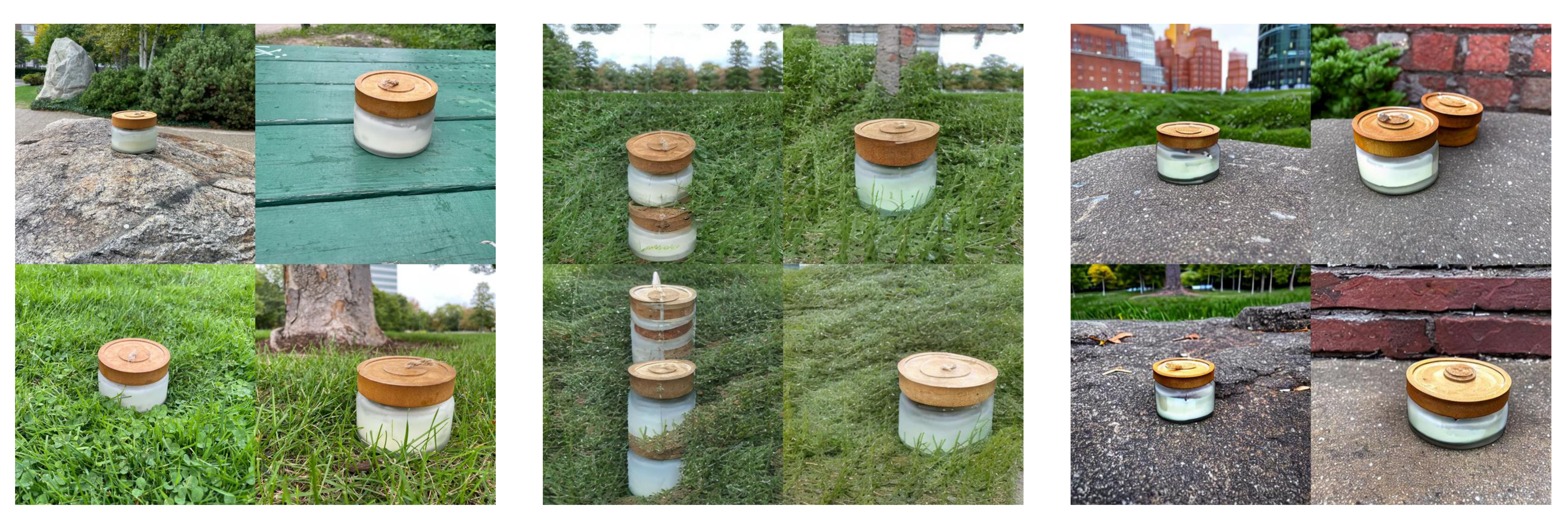}

\vspace{0.3cm}

\includegraphics[width=\linewidth]{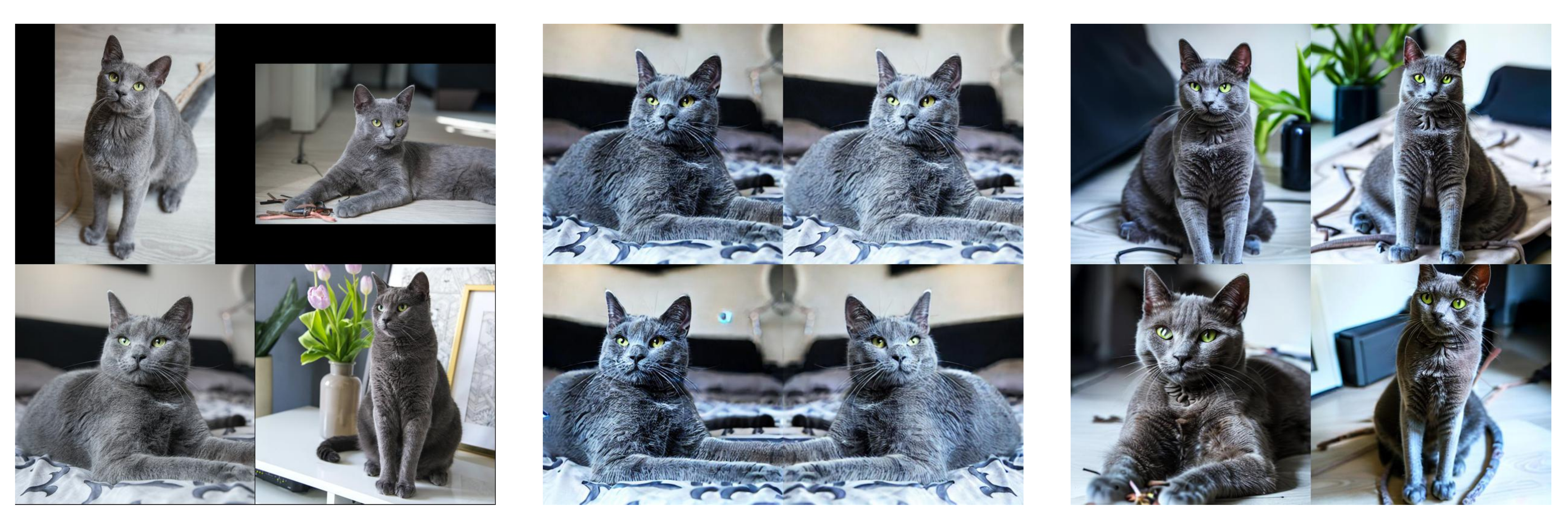}

\vspace{0.3cm}
\includegraphics[width=\linewidth]{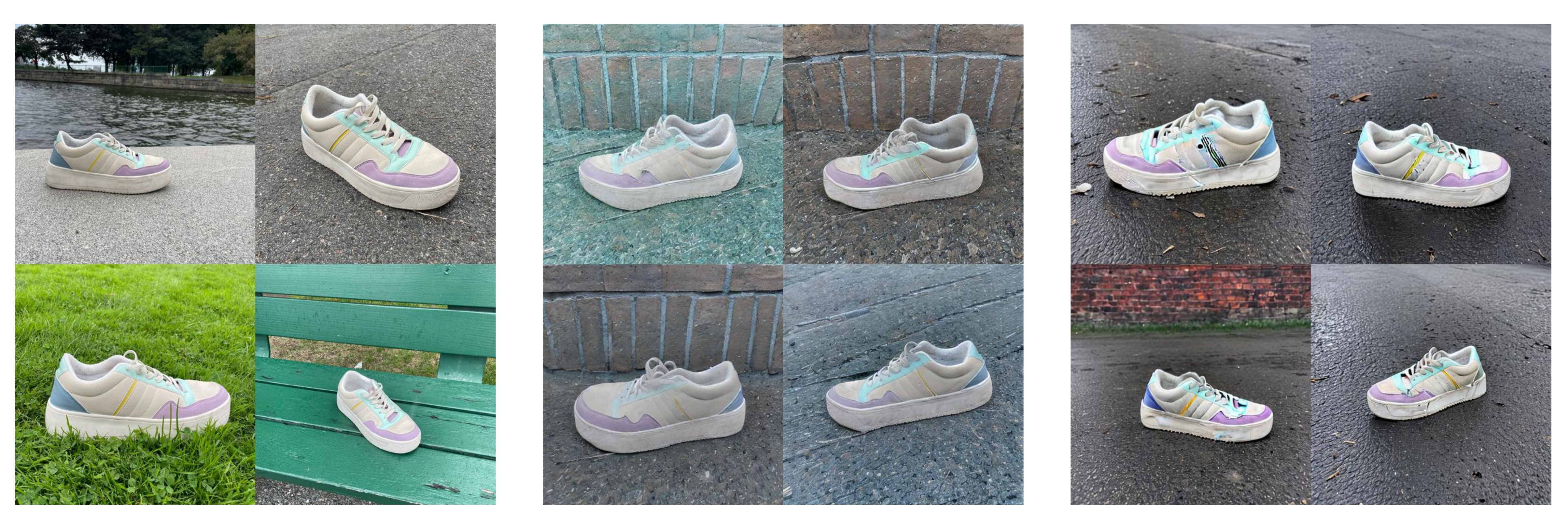}

\vspace{0.3cm}

\caption{\textbf{Example of DreamBooth}, train data (left), DreamBooth (middle), our method (right). All samples are randomly selected with the best training steps.}
\label{fig:dream example2}
\end{figure}

 \begin{figure}[H]
\includegraphics[width=\linewidth]{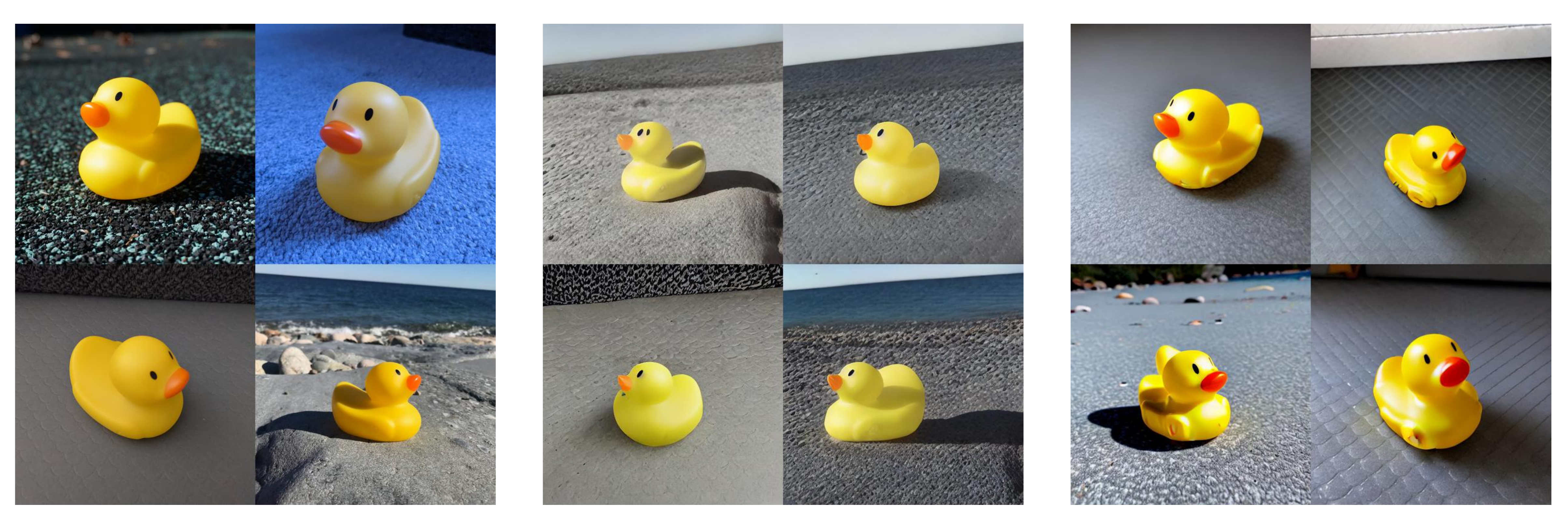}

\vspace{0.3cm}

\includegraphics[width=\linewidth]{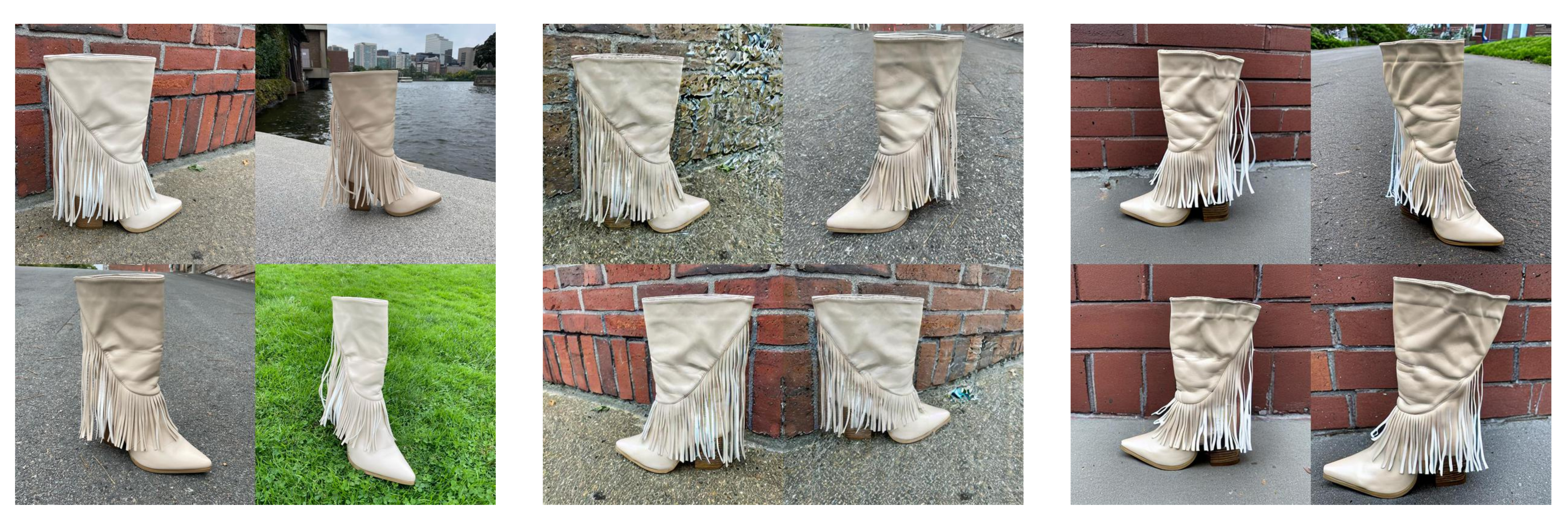}

\vspace{0.3cm}

\includegraphics[width=\linewidth]{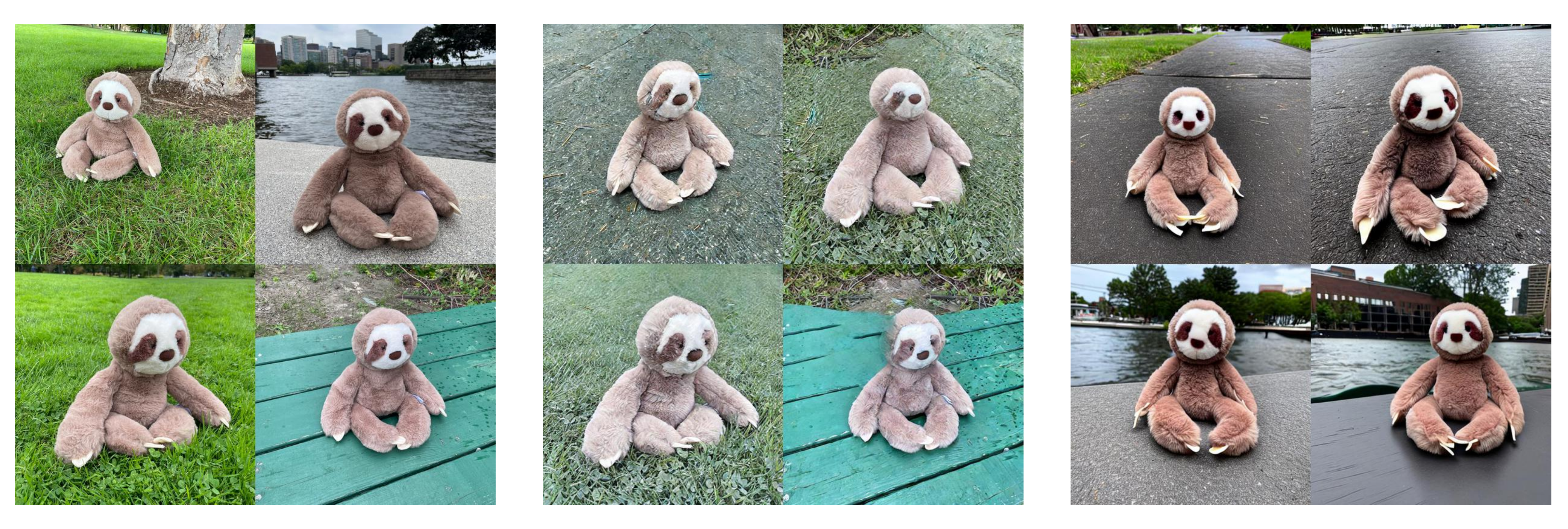}

\vspace{0.3cm}

\caption{\textbf{Example of DreamBooth}, train data (left), DreamBooth (middle), our method (right). All samples are randomly selected with the best training steps.}
\label{fig:dream example3}
\end{figure}

\textbf{Experiment detail.}
In the Figure-6 of the main paper, we find no significant performance changes after 1k steps of tuning in our main, so we use an averaged CLIP similarity between 1k and 2.5k training steps at an interval of 500 steps as the final metric.

In the Figure-7 of the main paper, due to methods convergence on different training steps, we use minimum FID in the training process as the final metric.

\end{document}